\documentclass{article} 
\usepackage{arxiv}

\usepackage{amsmath,amsfonts,bm}

\def\eqref#1{equation~\ref{#1}}

\def\1{\bm{1}}

\DeclareMathAlphabet{\mathsfit}{\encodingdefault}{\sfdefault}{m}{sl}
\SetMathAlphabet{\mathsfit}{bold}{\encodingdefault}{\sfdefault}{bx}{n}

\usepackage{hyperref}
\usepackage{url}
\usepackage{xcolor}
\usepackage[utf8]{inputenc} 
\usepackage[T1]{fontenc}    
\usepackage{hyperref}       
\usepackage{url}            
\usepackage{booktabs}       
\usepackage{amsfonts}       
\usepackage{nicefrac}       
\usepackage{microtype}      
\usepackage{lipsum}		
\usepackage{graphicx}
\usepackage{natbib}
\usepackage{doi}
\usepackage{graphicx}
\usepackage{subcaption}
\usepackage{multicol}
\usepackage{lipsum}
\usepackage{mwe}
\usepackage{tabularx}
\usepackage{amsmath}
\usepackage{booktabs}

\title{SAM Meets UAP: Attacking Segment Anything Model With Universal Adversarial Perturbation}
\author{
Dongshen Han \\
     Kyung Hee University \\
\And
Chaoning Zhang\thanks{You are welcome to contact the authors through chaoningzhang1990@gmail.com.} \\
 Kyung Hee University \\
\And
Sheng Zheng  \\
    Beijing Institute of Technology\\
\And
Chang Lu \\
    Kyung Hee University \\
\And
Yang Yang \\
University of Electronic Science and \\
Technology of China \\
\And
Heng Tao Shen \\
University of Electronic Science and\\
Technology of China \\
}

\begin{document}
\maketitle

\begin{abstract}
As Segment Anything Model (SAM) becomes a popular foundation model in computer vision, its adversarial robustness has become a concern that cannot be ignored. This works investigates whether it is possible to attack SAM with image-agnostic Universal Adversarial Perturbation (UAP). In other words, we seek a single perturbation that can fool the SAM to predict invalid masks for most (if not all) images. We demonstrate convetional image-centric attack framework is effective for image-independent attacks but fails for universal adversarial attack. To this end, we propose a novel perturbation-centric framework that results in a UAP generation method based on self-supervised contrastive learning (CL), where the UAP is set to the anchor sample and the positive sample is augmented from the UAP. The representations of negative samples are obtained from the image encoder in advance and saved in a memory bank. The effectiveness of our proposed CL-based UAP generation method is validated by both quantitative and qualitative results. On top of the ablation study to understand various components in our proposed method, we shed light on the roles of positive and negative samples in making the generated UAP effective for attacking SAM. 

\end{abstract}

\section{Introduction}

With an increasingly important role in driving groundbreaking innovations in AI, deep learning has gradually transitioned from training models for specific tasks to a general-purpose foundation model~\cite{bommasani2021opportunities}. For language foundation models like BERT~\cite{devlin2018bert} and GPT~\cite{radford2018improving, radford2019language}, have made significant breakthroughs in the natural language processing (NLP) area and contributed to the development of various generative AI~\cite{zhang2023asurvey}, including the text generation (ChatGPT~\cite{zhang2023ChatGPT}), text-to-image~\cite{zhang2023text} and text-to-speech~\cite{zhang2023audio}, text-to-3D~\cite{li2023generative}, etc. On top of the early successful attempts like masked encoder~\cite{zhang2022survey}, Meta research team has recently proposed a vision foundation model called Segment Anything Model (SAM)~\cite{kirillov2023segment}, which mimics GPT to control the output with prompts. Such a prompt-guided approach alleviates the need for finetuning and thus has impressive zero-shot transfer performance.  

After the release of \textit{Segment Anything} project, SAM has been widely used in various applications~\cite{zhang2023survey}, such as image editing~\cite{kevmo2023magiccopy} and object tracking~\cite{adamdad2023anthing, chen2023box}, etc. Therefore, it is critical to understand its robustness in various contexts. Early works~\cite{qiao2023robustness} have examined its generalization capabilities beyond natural images to medical images~\cite{zhang2023input} and camouflaged images~\cite{tang2023can}. Follow-up works have further evaluated its robustness under style transfer, common corruptions, patch occlusion and adversarial perturbation. Attack-SAM is a pioneering work to study how to attack SAM with adversarial examples, but it mainly focuses on image-independent attacks~\cite{zhang2023attack}. In other words, the generated perturbation can only be used for attacking the model for a specific image, which requires generating a new perturbation when the image changes. By contrast, a universal adversarial attack seeks a single perturbation (termed UAP) that causes the adversarial effect to all images and leads to wrong label predictions for most images~\cite{moosavi2017universal} in the context of image classification. With the image-agnostic property, the UAP can be generated beforehand and applied to any image for the attack purpose and thus is relatively more practical but also more challenging. Therefore, our work is devoted to studying whether it is possible to attack SAM with a UAP. 

Classical adversarial attack methods like DeepFool~\cite{moosavi2016deepfool} and PGD~\cite{madry2017towards} optimize the perturbation to make the output of the adversarial image different from that of the original clean image. The classical UAP algorithm introduced in~\cite{moosavi2017universal} is based on DeepFool and thus follows such an image-centric approach. This requires access to the original training data and thus FFF~\cite{Mopuri2017datafree} studies PGD-based approaches for generating data-free UAP~\cite{Mopuri2017datafree} with a relatively weaker attack performance. Prior works~\cite{qiao2023robustness,zhang2023attack} show that such an image-centric approach is also effective for attacking SAM, but the investigation is limited to image-independent attacks. A major difference in generating UAP lies in changing the to-be-attacked training image in every iteration to avoid over-fitting to any specific image. We follow this practice to extend Attack-SAM from image-independent attacks to universal attacks, however, such a preliminary investigation leads to unsatisfactory performance. This is attributed to the change of optimization target from one image to another in the image-centric approach. To this end, this work proposes a new perturbation-centric attack method, by shifting the goal from directly attacking images to seeking augmentation-invariant property of UAP. Specifically, we optimize the UAP in the CL method where the UAP is chosen as the anchor sample. The positive sample is chosen by augmenting the anchor sample, while random natural images are chosen as the negative samples. 

For the proposed CL-based UAP generation method, we experiment with various forms of augmentations to generate a positive sample and find that augmenting the UAP by adding natural images yields the most effective UAP for universal adversarial attack. Beyond quantitative verification, we also show visualize the attack performance of the generated UAP under both both point and box prompts. We have an intriguing observation that the predicted mask gets invalid under both types of prompts: getting smaller under point prompts and getting larger under box prompts. Moreover, we present a discussion to shed light on why our generated UAP is effective by analyzing different pairs of inputs for the encoded feature representations. It helps us understand the roles of positive samples and negative samples in our CL-based UAP method for crafting an effective UAP to attack SAM.

\section{Related works}

\textbf{Segment Anything Model (SAM).} SAM is a recent advancement in the field of computer vision that has garnered significant attention \cite{ma2023segment, zhang2023input, tang2023can, han2023segment, shen2023anything, kang2022any}. Unlike traditional deep learning recognition models focusing solely on label prediction, SAM performs mask prediction tasks using prompts. This innovative approach allows SAM to generate object masks for a wide range of objects, showcasing its remarkable zero-shot transition performance. Researchers have explored the reliability of SAM by investigating its susceptibility to adversarial attacks and manipulating label predictions. Furthermore, SAM has been extensively utilized in various applications, including medical imaging \cite{ma2023segment, zhang2023input}, and camouflaged object detection \cite{tang2023can}. It has also been combined with other models and techniques to enhance its utility, such as combining with Grounding DINO for text-based object detection and segmentation \cite{GroundedSegmentAnything2023} and integrating with BLIP or CLIP for label prediction \cite{chen2023semantic, park2023segment, li2022blip, radford2021learning}. SAM has found applications in image editing \cite{rombach2022high}, inpainting \cite{yu2023inpaint}, and object tracking in videos \cite{yang2023track, z-x-yang_2023}. More recently, MobileSAM~\cite{zhang2023faster}, which is significantly smaller and faster than the original SAM, realizes lightweight SAM on mobile devices by decoupled knowledge distillation. With the advent of MobileSAM, it is expected more and more SAM-related applications will emerge, especially in the computation-constrained edge devices. This yields a need to understand how SAM works, for which zhang~\cite{zhang2023understanding} performs a pioneering study and shows that SAM is biased towards texture rather than shape. Moreover, multiple works~\cite{qiao2023robustness,zhang2023attacksam} have shown that SAM is vulnerable to the attack of adversarial examples. Our work also investigates the adversarial robustness of SAM, but differentiates by focusing on universal adversarial attack. 

\textbf{Universal Adversarial Attack.} Universal adversarial perturbation (UAP) has been first introduced in~\cite{moosavi2017universal} to fool the deep classification model by making wrong label predictions for most images. Unlike the vanilla universal attack by the projected algorithm to generate the perturbations, the SV-UAP~\cite{khrulkov2018art} adopts singular vectors to craft UAPs, where the method is data-efficient with only 64 images used to iteratively craft the perturbations. Inspired by the Generative Adversarial Networks (GAN), NAG~\cite{mopuri2018nag} and GAP~\cite{perolat2018playing} focus on obtaining the distribution of UAPs. To compute the UAPs, these approaches use a subset of the training dataset, however, the attacker might be limited in accessing the training data. Therefore, multiple works explore data-free to generate UAPs. FFF~\cite{mopuri2017fast} is pioneering to propose a data-independent approach to generate the UAPs, adopting fooling the features learned at multiple layers. GD-UAP~\cite{mopuri2018generalizable} can generate universal perturbations and transfer to multiple vision tasks. Class-discriminative UAP has been investigated in~\cite{zhang2019cd-uap,benz2020double} to fool the model for a subsect of classes while minimizing the adversarial effect on other classes of images. They opt to train the UAP with Adam Optimizer~\cite{kingma2014adam} instead of adopting sign-based PGD algorithms~\cite{goodfellow2014explaining,madry2017towards}, and such a practice has also been adopted in ~\cite{zhang2020understanding,zhang2021data}. In contrast to prior works adopting image-centric DeepFool or PGD to optimize the UAP, our work proposes a perturbation-centric framework with a new UAP generation method based on contrastive learning.   

\textbf{Self-supervised Contrastive Learning (CL).} With the goal of learning augmentation-invariant representation, for which CL is a miltstone development of unsupervised learning~\cite{Schroff2015FaceNetAU,wang2015unsupervised,sohn2016improved,misra2016shuffle,Federici2020Learning}. CL consists of positive pair and negative pairs. Unlike the negative pairs, the positive pair are obtained from the same image but differ in augmentation to ensure they have similar semantic information. Earlu works on CL have adopted margin-based contrastive losses~\cite{hadsell2006dimensionality,wang2015unsupervised,hermans2017defense}, and NCE-like loss\cite{wu2018unsupervised,oord2018representation} has later emerged to become the de facto standard loss in CL. For example, classical CL methods like SimCLR~\cite{chen2020simple} and MoCo families~\cite{he2020momentum,chen2020mocov2} adopt the InfoNCE loss which combines mutual information and NCE. Specifically, it maximizes the mutual information between the representation of different views in the same scene. 

\section{Background and Problem Formulation}

\subsection{Prompt-guided Image Segmentation}
\label{sec:preliminaries}
Segment Anything Model (SAM) consists of three components: an image encoder, a prompt encoder, and a lightweight mask decoder. The image encoder adopts the MAE~\cite{he2022masked} pre-trained Vision Transformer (ViT), which generates the image representation in the latent space. The prompt encoder utilizes positional embeddings to represent the prompt (like points and boxes). The decoder takes the outputs of image and prompt encoders as the inputs and predicts a valid mask to segment the object of interest. In contrast to classical semantic segmentation performing pixel-wise label prediction, the SAM generates a label-free mask. With $x$ and $p$ denoting the image and prompt, respectively, we formalize the mask prediction of SAM as follows: 

\begin{equation}
y = SAM(x, p; \theta),
\label{eq:sam_forward}
\end{equation}
\noindent where $\theta$ represents the parameter of SAM. 
Given a image $x\in\mathbb{R}^{H\times W\times C}$, the shape of y is $\mathbb{R}^{H\times W}$. We set the $x_{ij}$ as the pixel values at the image x with the coordinates i and j. $x_{ij}$ belongs to the masked area if the pixel value $y_{ij}$ is larger than the threshold of zero. 
 
\subsection{Universal Adversarial Attack on SAM}

Here, we formalize the task of universal adversarial attack on SAM. Let $\mu$ denote the distribution of images in $\mathbb{R}^{H\times W\times C}$. In the image recognition tasks, the adversary goal is to fool the model to predict wrong labels. Universal adversarial attack, under the assumption that the predicted labels of clean images are the correct ones, seeks a \textit{single} perturbation vector $v\in\mathbb{R}^{H\times W\times C}$ termed UAP to cause label changes for \textit{most} images~\cite{moosavi2017universal}. In other words, it aims to maximize the adversarial effect of the UAP in terms of the fooling rate, the ratio of images whose predicted label changes after adding the UAP~\cite{moosavi2017universal}. In the context of SAM, the predicted outputs are masks instead of labels and thus the attack goal is to cause mask changes. We follow Attack-SAM to adopt the widely used Intersection over Union (IoU) in image segmentation to evaluate such mask changes. The mIoU calculates the mean IoU for $N$ pairs of clean mask $Mask_{clean}$ and adversarial mask $Mask_{adv}$ shown in Equation~\ref{eq:miou}. 

\begin{equation}
mIoU = \frac{1}{N}\sum_{n=1}^{N} IoU(Mask^{(n)}_{clean}, Mask^{(n)}_{adv}),
\label{eq:miou}
\end{equation}

where all the adversarial masks $Mask_{adv}$ are generated for all $N$ images by a single UAP. The goal of universal adversarial attack on SAM is to seek such a single perturbation $v$ to decrease the mIoU defined in Eq.~\ref{eq:miou} as much as possible. The UAP $v$ is bounded by a $l_{p}$ norm, which is set to $l_\infty$ norm  conventions in prior works on SAM~\cite{moosavi2017universal,moosavi2017analysis}.

\paragraph{Implementation details.} Considering the image-agnostic property, $N$ in Eq.~\ref{eq:miou} needs to be larger than 1 and is set to 100 in this work. For the prompts, we randomly choose point prompts unless specified otherwise. Specifically, we randomly select 100 test images from the SA-1B dataset~\cite{kirillov2023segment} for evaluating the generated UAP. Note that the test images cannot be used for generating the UAP. Following the existing works on the universal adversarial attacks in computer vision, we use $10/255$ as the maximum limit for the perturbation. In other words, the allowed maximum change on each pixel can be no bigger than $10/255$. 

\section{Method}

\subsection{Existing Image-Centric Attack Framework}
For the task of adversarial attack, the goal is to make the deep model predict invalid output after adding a small perturbation on the input image. Therefore, numerous attack methods, including classical DeepFool~\cite{moosavi2016deepfool} and PGD~\cite{madry2017towards}, optimize such an adversarial perturbation to make the output of adversarial image different from that of its clean image. Such an image-centric approach consists of two steps. First, it predicts the output of clean image $y_{clean}$ and saves it as the ground-truth\footnote{the ground-truth output might be given at first in some cases, where this step can be skipped.}. Second, the perturbation in the adversarial image is optimized to make $y_{adv}$ different from the ground-truth $y_{clean}$. 

Universal adversarial attack requires the perturbation to be effective at random unseen images. Therefore, the to-be-attacked training image needs to be changed in every iteration of the optimization process to avoid over-fitting on any single training image. Such an image-centric approach has been adopted in~\cite{zhang2023attacksam} to demonstrate successful image-independent attacks, and we have adapted it to image-agnostic, \textit{universal} adversarial attacks. The results in Table~\ref{tab:image_centric} show that the generated UAP performs much better than random uniform noise sampled between $-10/255$ and $10/255$. Nonetheless, the value of mIoU ($59.50\%$) is still quite high, demonstrating that the UAP is not sufficiently effective for causing mask changes. We also experiment with not changing the to-be-attacked image, which fixes the same optimization goal and results in a successful image-dependent attack with a mIoU of 0.0\%. This suggests that a successful attack in SAM requires a consistent optimization target (like attacking a single image). However, such success is limited to image-dependent attacks due to overfitting and cannot be generalized to unseen test images. 

\begin{table}[!htbp]
  \caption{mIoU (\%) results of Image-centric attack by uniform noise and adversarial examples. Image-agnostic indicates the universal setup to attack unseen images.}
  \label{tab:image_centric}
  \centering
  \begin{tabular}{lcccc}
    \toprule
     Input & Image-dependent & Image-agnostic \\
    \midrule 
    Uniform noise & 86.97 & 86.97\\ 
    Adversarial attack & 0.0 & 59.50  \\
  \bottomrule
\end{tabular}
\end{table}

\subsection{Proposed Perturbation-Centric Attack Framework}
\begin{figure}[!htbp]
     \centering
     \begin{minipage}[b]{1.0\textwidth}
         \centering
         \includegraphics[width=\textwidth]{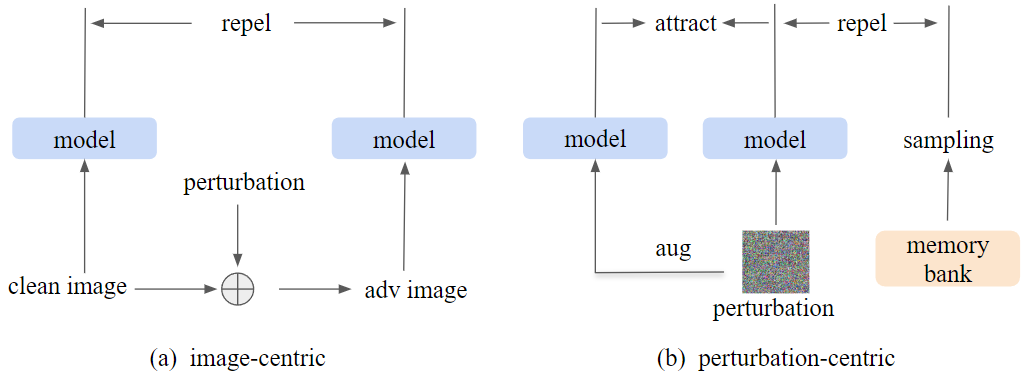}
     \end{minipage}
        \caption{Difference between image-centric (left) and perturbation-centric (right) attack frameworks.}
    \label{fig:cl_framework}
\end{figure}

The above image-centric method is suitable for image-independent attack on SAM but fails for universal attack. The image-centric method is in essence a supervised approach where $y_{clean}$ plays the role of ground-truth and the added perturbation is optimized to make $y_{adv}$ far from $y_{clean}$. Such a supervised approach inevitably causes a dramatic change to the optimization goal when the training image is changed at every iteration. In other words, the failure of image-centric approach for universal attack is conjectured to be the inconsistent optimization goal caused by the change of training image at every iteration. Therefore, we shift the perspective from image to perturbation, which results in our proposed perturbation-centric method. Specifically, in contrast to the predicted masks of the clean and adversarial images, we focus on the independent features of the UAP, which is motivated by perceiving the UAP as an independent input considering its image-agnostic property. How to optimize the UAP in such a perturbation-centric approach, however, is a non-trivial task. It cannot be straightforwardly optimized in a supervised manner as in the image-centric method. To this end, we turn to a widely used self-supervised approach known as \textit{Contrastive Learning (CL)}. The difference between image-centric and perturbation-centric framework is summarized in Figure~\ref{fig:cl_framework}. 

\textbf{CL-based UAP Genration Method.} Outperforming its supervised counterpart, self-supervised learning has become a dominant approach for pre-training an backbone encoder, where CL is a widely adopted method. In the classical CL , there are three types of samples: anchor sample, positive sample, and negative sample. The anchor sample is the sample of interest, while the positive sample is augmented from the anchor sample. Other random images are chosen as the negative samples, and we adopt the same practice in our CL-based UAP generation method. What makes it different from the classical CL method lies in the choice of anchor sample. Specifically, the UAP ($v$) is chosen as the anchor sample because it is the input of interest in this context. For the positive sample, we obtain it by augmenting the anchor sample UAP, which will be discussed in detail.  The NCE-like loss (often termed InfoNCE loss) has been independently introduced in multiple works and constitutes as the de-facto standard loss for CL. Following~\cite{he2020momentum}, we denote the encoded features of the anchor sample, positive sample, and negative sample with $q$, $k_{+}$ and $k_{-}$, respectively. Note that the encoded features are often L2 normalized to remove scale ambiguity, based on which the InfoNCE loss adopted in the CL-based UAP generation method is shown as follows:
\begin{equation}
L_{infonce} = -log\frac{exp(q\cdot k_{+}/ \tau)}{exp(q\cdot k_{+}/ \tau) + \sum_{i=1}^{K}exp(q\cdot k^{i}_{-}/ \tau)},
\label{eq:infonce}
\end{equation}
where $\tau$ represents the temperature controlling the hardness-aware property and thus has an implicit influence on the size of negative samples~\cite{Wang_2021_CVPR,zhang2022dual}. A large negative sample size is required to better sample the high-dimensional visual space~\cite{he2020momentum}. We follow prior works to save the encoded features of negative samples in a list termed as memory bank~\cite{wu2018unsupervised} or dictionary~\cite{he2020momentum}. Since the to-be-attacked SAM encoder does not change during the optimization of UAP, the list does not need to be updated as in classical CL method~\cite{wu2018unsupervised,he2020momentum}. In other words, the $k_{-}$ in Eq~\ref{eq:infonce} can be generated once and then saved for sampling during the optimization of UAP. 

In the classical CL method, augmentation is applied to ensure augmentation-invariant property for the encoder learning meaningful representations. In our CL-based UAP method, augmentation is also essential for making the generated UAP cause augmentation-invariant feature response on the encoder. This yields two intertwined questions: (1) how should we choose such augmentation for making the UAP effective? (2) why does such augmentation-invariant property makes the UAP effective? The following section performs an empirical study to shed insight on these two intertwined questions.

\section{Experimental Results and Analysis}

\subsection{Towards Finding Effective Augmentation}

\textbf{Preliminary investigation.} In the classical CL method, there are mainly two types of augmentations~\cite{chen2020simple}. The first type involves spatial transformation like crop/resize and cutout. The second type involves no spatial transformation but causes appearance change by adding low-frequency content (like color shift) or high-frequency content (like noise). We experiment with both types of augmentation and the results are shown in Table~\ref{tab:augmentation}. 
We observe that the mIoU values with augmentation crop/resize and cutout consistently remain high, at 85.11$\%$ and 75.48$\%$, respectively. It suggests that the spatial transformation is not an effective augmentation type in our UAP generation method. For the second type of adding content, adding uniform noise is also not effective with a mIoU value of 81.14$\%$. By contrast, the augmentation of color shift yields a mIoU of 61.64$\%$, which is comparable to that of the image-centric method (see 59.5$\%$ in Table \ref{tab:image_centric}). 

\begin{table}[!htbp]
  \caption{Comparison of different augmentations. The Crop size is 200$\times$200 out of 1024$\times$1024, cutout size is 200$\times$200. The uniform noise and color shift are ranged from 0 to 255. Adding natural images achieves significantly better performance than other augmentations. 
  }
  \label{tab:augmentation}
  \centering
  \begin{tabular}{lcccccc}
    \toprule
        Augmentation type & mIoU ($\downarrow$)\\
    \midrule 
    Crop/Resize & 85.11 \\ 
    Cutout & 75.48 \\
    Uniform noise&81.14\\ 
    Color shift & 61.64\\
    \midrule
    Adding natural images & 15.01\\ 
  \bottomrule
\end{tabular}
\end{table}

\textbf{From color shift to natural images.} Our preliminary investigation suggests that color shift is the most effective augmentation among those we investigate. We believe that this might be connected to how the generated UAP is applied to attack the model in practice. Since UAP is directly added to the images without spatial transformation, which explains why spatial transformation is less effective. Moreover, natural images have the property of being locally smooth and thus mainly contain low-frequency content, which justifies why the color shift is relatively more effective than adding noise. Motivated by the above interpretations, we conjecture that replacing the color shift images with random natural images for additive augmentation is beneficial for higher attack performance, which is supported by the results in Table~\ref{tab:augmentation}. Here, for simplicity, the weight of the augmented natural images is set to 1. However, it can be set to values different from 1 (see the ablation study results in Figure~\ref{fig:result_different_weights}).

\subsection{Qualitative Results}
\begin{figure*}[!htbp]
    \centering
     \begin{minipage}[t]{0.85\textwidth}
         \centering
         \includegraphics[width=\textwidth]{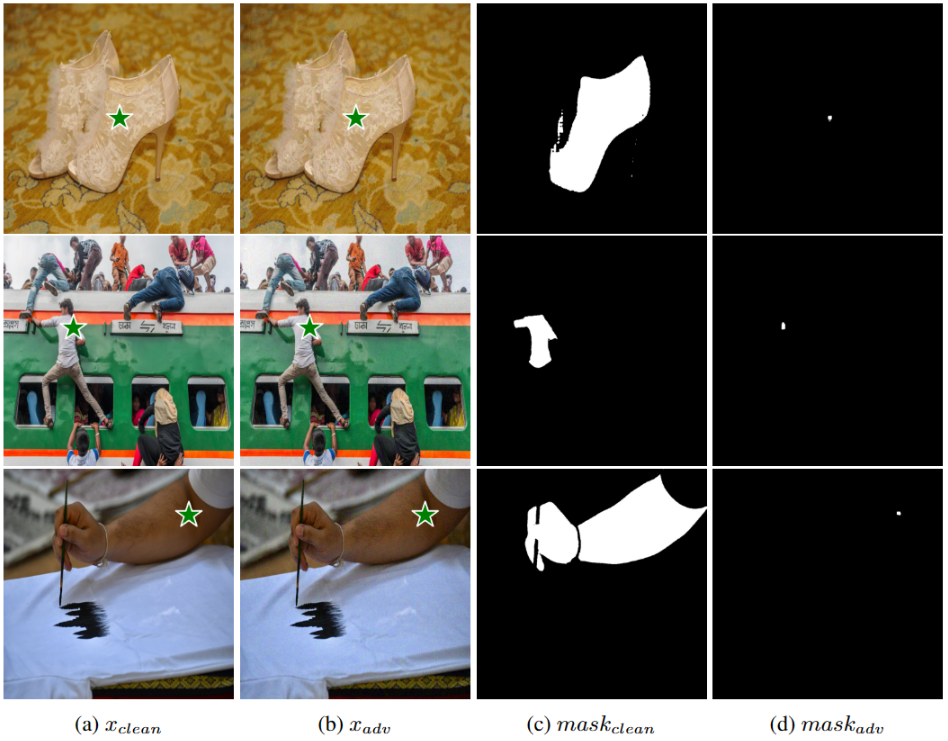}
     \end{minipage} 
        \caption{Qualitative results under point prompts. Column (a) and (b) shows the clean and adversarial images with the point prompt marked in a green star, with their predicted masks shown in column (c) and (d), respectively. The UAP makes the mask invalid by removing it (or making it smaller).
        }
    \label{fig:point}
\end{figure*}

\begin{figure*}[!htbp]
    \centering
     \begin{minipage}[t]{0.85\textwidth}
         \centering
         \includegraphics[width=\textwidth]{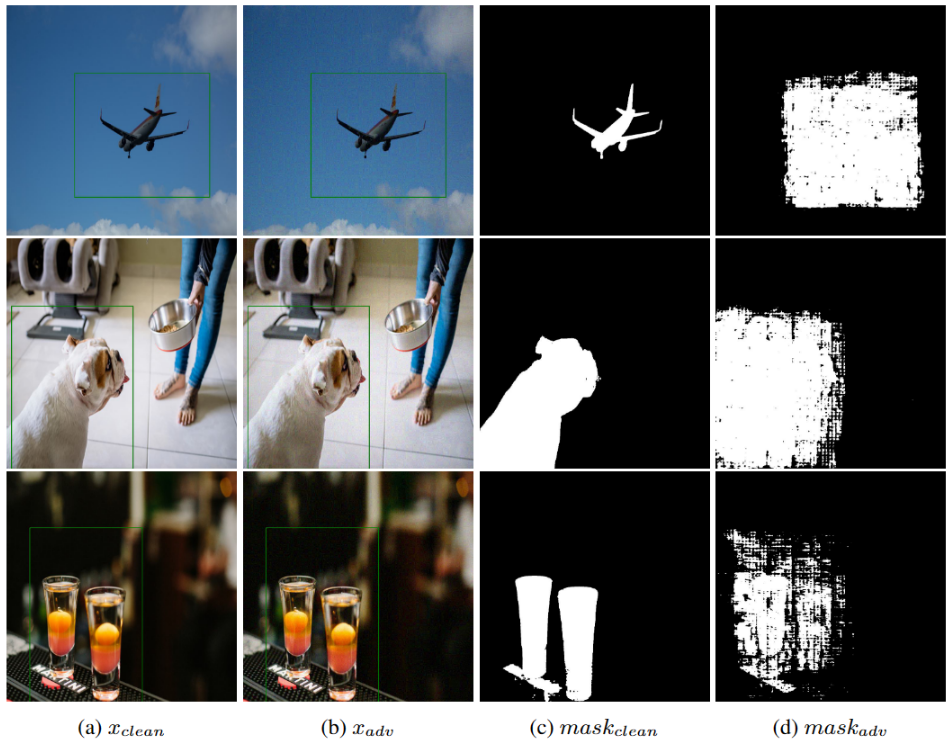}
     \end{minipage} 
        \caption{Qualitative results under box prompts. Column (a) and (b) refers to the clean and adversarial images with the box prompt marked with green lines, with their predicted masks shown in column (c) and (d), respectively. The UAP makes the mask invalid by making it larger and blurry.  
        }
    \label{fig:box}
\end{figure*}

It is worth highlighting that our generated UAP has one hidden merit it can generalize to all prompts because the UAP is optimized only on the SAM encoder. In other words, it is truly universal in the sense of being both image-agnostic and prompt-agnostic. In the above, we only report the quantitative results under random point prompts. Here, for qualitative results, we visualize the attack performance under both point prompts and box prompts, with results shown in Figure~\ref{fig:point} and Figure~\ref{fig:box}, respectively. We find that the single UAP causes the model to produce invalid masks for both types of prompts but with an intriguing distinction. Specifically, under the point prompts, the predicted mask region gets smaller with a boundary close to the chosen point prompt. Under the box prompt, however, the predicted mask gets larger than the original mask. We have no clear explanation for this intriguing phenomenon. A possible explanation is that the UAP tends to cause the predicted output to have similar values, \textit{i.e.} causing confusion between the original masked regions and unmasked regions. For the point prompt, the unmasked region tends to be much larger than that of the masked region and thus the predicted mask gets smaller after UAP. By contrast, the box prompts tends to predict a mask inside the box, and thus tends to make the predicted mask boundary get larger and vauge. Note that we can still observe the glass mask in the third row of Figure~\ref{fig:box}, but the mask boundary gets blurred. 

\subsection{Ablation Study}

\begin{figure}[!htbp]
     \centering
     \begin{minipage}[b]{0.85\textwidth}
         \centering
         \includegraphics[width=\textwidth]{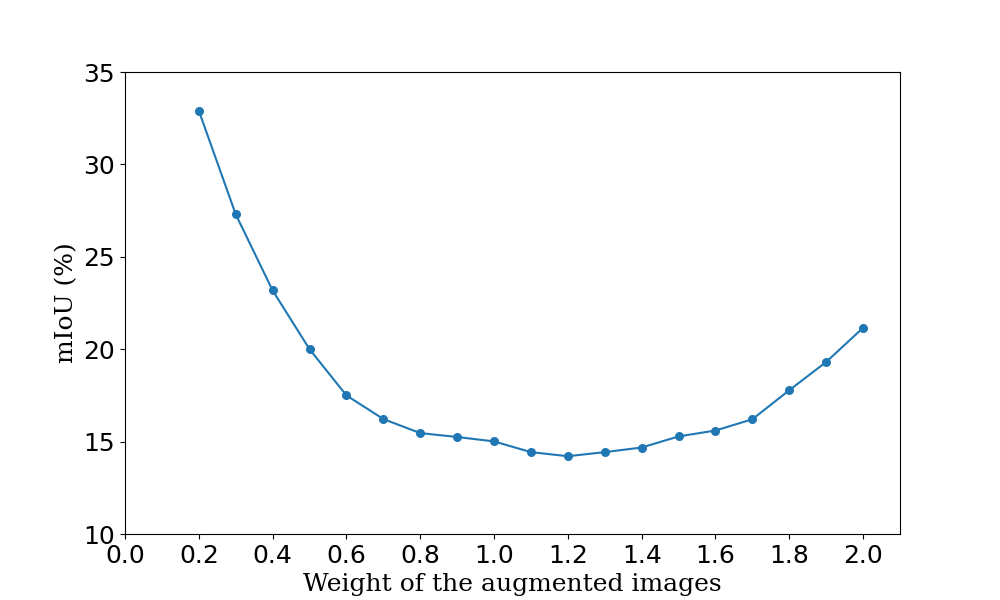}
     \end{minipage}
        \caption{The mIoU (\%) results for different weights of the augmented images.  
        }
    \label{fig:result_different_weights}
\end{figure}

\textbf{Weight of Augmented Images.} Here, we first conduct an ablation study on the weight of the augmented images. The results are shown in Figure~\ref{fig:result_different_weights}. We observe that the mIoU value decreases first increases and then decreases when the weight value is increased from 0.2 to 2 with an interval of 0.1. The strongest attack performance with the mIoU value of 14.21 appears when the weight is set to 1.2. Overall the mIoU value stays low for a relatively wide range of augmentation weight, suggesting our proposed method is moderately sensitive to the choice of augmentaiton weight. 

\textbf{Size of Negative Sample.} 
For negative samples in contrastive learning, unlike the positive samples that aim to attract the anchor, our objective is to create a repelling effect on the anchor. This enables the anchor to more effectively foucs on independent features by being drawn towards the positive samples. To accomplish this, it is essential to incorporate a diverse set of negative sample representations, thus avoiding repetitive generation. Therefore, we implement the memory bank mechanism, as do in prior work. We use various sample numbers ( 1, 2, 5, 10, 20, 50, 100) as our memory bank. As shown in Table 3, we observe a significant increase in universal attack performance as the number of samples increases. This indicates that augmenting diverse negative sample representations through the memory bank is beneficial for UAP training. To further augment diverse negative sample representations,

\begin{table}[!htbp]
  \caption{The mIoU (\%) results on different negative samples $N$.}
  \label{tab:uap_attack}
  \centering
  \begin{tabular}{lccccccc}
    \toprule
        N & 1 & 2& 5 &10 &20 &50&100 \\
    \midrule 
    mIoU ($\downarrow$) & 38.91 & 30.71 & 24.83 & 19.88 & 17.63 & 15.92 &15.01\\ 
    
  \bottomrule
\end{tabular}
\end{table}

\textbf{Temperature.} Temperature is widely known to have a large influence on the performance of CL method~\cite{Wang_2021_CVPR,zhang2022dual}. The influence of temperature in our CL-based UAP method is shown in Table ~\ref{tab:iou_different_temperatures}. By default, the temperature is set to 0.1 in this work. We observe that the temperature significantly decreases when the temperature is set to a large value. The reason is that a smaller temperature causes more weight on spent on those hard negative samples~\cite{Wang_2021_CVPR,zhang2022dual}. As revealed in~\cite{zhang2022dual}, a small temperature is equivalent to choosing a small negative sample size. Therefore, it is well expected that the attack performance decreases when the temperature is set to a sufficiently small value because the a relatively large negative sample size is required for CL. Unlike classical CL, a relatively large temperature does not cause a performance drop.

\begin{table}[!htbp]
  \caption{The mIoU (\%) results on different InfoNCE temperatures.}
  \label{tab:iou_different_temperatures}
  \centering
  \begin{tabular}{lccccccccc}
    \toprule
        Temperature & 0.005 & 0.01 & 0.05 & 0.1 & 0.5 & 1\\
    \midrule
    mIoU ($\downarrow$) & 64.61 & 60.58 &22.78&  15.01 & 13.28 & 13.48 \\
    \bottomrule
\end{tabular}
\end{table}

\subsection{Discussion}
To shed more light on why the generated UAP is effective in attacking unseen images, we analyze the cosine similarity of different pairs of inputs for the encoded feature representations, and the results are shown in Table~\ref{tab:analysis}. The positive sample pairs have a much higher cosine similarity than that of the negative sample pairs, which aligns with our training objective in Eq~\ref{eq:infonce}. The cosine similarity between pairs of adversarial images and its clean images is higher than that of the negative sample pairs, which is excepted because the adversarial image consists of a random natural image and the UAP. The fact that the cosine similarity between positive sample pairs is very high (0.87) suggests that the UAP has independent features and it can be robust against the augmentation of image addition, which aligns with the finding in~\cite{zhang2020understanding}. This partly explains why the cosine similarity between pairs of clean images and adversarial images is relatively low (0.40), causing a successful universal attack. In other words, how the generated UAP attacks the model does not intend to identify the vulnerable spots in the clean images to fool the model as suggested in~\cite{moosavi2017universal,moosavi2017analysis}, but instead form its own augmentation-invariant features. 

For the role of negative samples in Eq~\ref{eq:infonce}, we find that it is can be at least partially attributed to the the existence of common feature representations regardless of the image inputs for the image encoder, which is supported by a Cosine similarity value of 0.55 higher than zero for pairs of random images. With a list of negative samples in Eq~\ref{eq:infonce}, the UAP is expected to be optimized to offset such common features, thus causing adversarial effects. This interpretation is partially supported by the comparison between 0.40 and 0.55 in Table~\ref{tab:analysis}. 

Overall, the success of Eq~\ref{eq:infonce} for generating an effective UAP can be interpreted as follows: the role of the positive sample is to make the UAP have independent features that are robust against the disturbance of natural image, while the role of negative images facilitates the UAP to find more effective directions to cause adversarial effects by partially canceling out the common feature representations in the image encoder. We leave further detailed analysis to future works. 

\begin{table}[!htbp]
  \caption{Cosine similarity analysis with different pairs of inputs.}
  \label{tab:analysis}
  \centering
  \begin{tabular}{lcc}
    \toprule
    Input pairs    & Cosine similarity \\
    \midrule 
   Positive sample pairs (UAP and augmented UAP) & 0.87 \\
   Negative sample pairs (UAP and random image) &0.34\\ 
   Pairs of adversarial image and its clean image & 0.40\\
   Pairs of two random images & 0.55\\
  \bottomrule
\end{tabular}
\end{table}

\section{Conclusion}
Our work is the first to study how to perform adversarial attack SAM with a single UAP. We demonstrate that existing image-centric attack framework is effective for image-dependent attacks but fails to achieve satisfactory performance for universal adversarial attacks. We propose a perturbation-centric attack framework resulting in a new generation method based on contrastive learning, where the UAP is set to the anchor sample. We experiment with various forms of augmentations and find that augmenting the UAP by adding a natural image yields the most effective UAP among all augmentations we have explored. The effectiveness of our proposed method has been verified with both qualitative and quantitative results. Moreover, we have presented and analyzed different pairs of inputs for the encoded feature representations, which shed light on the roles of positive samples and negative samples in our CL-based UAP method for crafting an effective UAP to attack SAM.   

\bibliographystyle{unsrtnat}
\bibliography{bib_mixed,bib_local,bib_sam,bib_targeted}

\begin{thebibliography}{66}
\providecommand{\natexlab}[1]{#1}
\providecommand{\url}[1]{\texttt{#1}}
\expandafter\ifx\csname urlstyle\endcsname\relax
  \providecommand{\doi}[1]{doi: #1}\else
  \providecommand{\doi}{doi: \begingroup \urlstyle{rm}\Url}\fi

\bibitem[Bommasani et~al.(2021)Bommasani, Hudson, Adeli, Altman, Arora, von
  Arx, Bernstein, Bohg, Bosselut, Brunskill,
  et~al.]{bommasani2021opportunities}
Rishi Bommasani, Drew~A Hudson, Ehsan Adeli, Russ Altman, Simran Arora, Sydney
  von Arx, Michael~S Bernstein, Jeannette Bohg, Antoine Bosselut, Emma
  Brunskill, et~al.
\newblock On the opportunities and risks of foundation models.
\newblock \emph{arXiv preprint arXiv:2108.07258}, 2021.

\bibitem[Devlin et~al.(2018)Devlin, Chang, Lee, and Toutanova]{devlin2018bert}
Jacob Devlin, Ming-Wei Chang, Kenton Lee, and Kristina Toutanova.
\newblock Bert: Pre-training of deep bidirectional transformers for language
  understanding.
\newblock \emph{arXiv preprint arXiv:1810.04805}, 2018.

\bibitem[Radford et~al.(2018)Radford, Narasimhan, Salimans, Sutskever,
  et~al.]{radford2018improving}
Alec Radford, Karthik Narasimhan, Tim Salimans, Ilya Sutskever, et~al.
\newblock Improving language understanding by generative pre-training.
\newblock 2018.

\bibitem[Radford et~al.(2019)Radford, Wu, Child, Luan, Amodei, Sutskever,
  et~al.]{radford2019language}
Alec Radford, Jeffrey Wu, Rewon Child, David Luan, Dario Amodei, Ilya
  Sutskever, et~al.
\newblock Language models are unsupervised multitask learners.
\newblock \emph{OpenAI blog}, 2019.

\bibitem[Zhang et~al.(2023{\natexlab{a}})Zhang, Zheng, Li, Qiao, Kang, Shan,
  Zhang, Qin, Rameau, Bae, et~al.]{zhang2023asurvey}
Chaoning Zhang, Sheng Zheng, Chenghao Li, Yu~Qiao, Taegoo Kang, Xinru Shan,
  Chenshuang Zhang, Caiyan Qin, Francois Rameau, Sung-Ho Bae, et~al.
\newblock Asurvey on segment anything model (sam): Vision foundation model
  meets prompt engineering.
\newblock 2023{\natexlab{a}}.

\bibitem[Zhang et~al.(2023{\natexlab{b}})Zhang, Zhang, Li, Qiao, Zheng, Dam,
  Zhang, Kim, Kim, Choi, et~al.]{zhang2023ChatGPT}
Chaoning Zhang, Chenshuang Zhang, Chenghao Li, Yu~Qiao, Sheng Zheng,
  Sumit~Kumar Dam, Mengchun Zhang, Jung~Uk Kim, Seong~Tae Kim, Jinwoo Choi,
  et~al.
\newblock One small step for generative ai, one giant leap for agi: A complete
  survey on chatgpt in aigc era.
\newblock \emph{arXiv preprint arXiv:2304.06488}, 2023{\natexlab{b}}.

\bibitem[Zhang et~al.(2023{\natexlab{c}})Zhang, Zhang, Zhang, and
  Kweon]{zhang2023text}
Chenshuang Zhang, Chaoning Zhang, Mengchun Zhang, and In~So Kweon.
\newblock Text-to-image diffusion models in generative ai: A survey.
\newblock \emph{arXiv preprint arXiv:2303.07909}, 2023{\natexlab{c}}.

\bibitem[Zhang et~al.(2023{\natexlab{d}})Zhang, Zhang, Zheng, Zhang, Qamar,
  Bae, and Kweon]{zhang2023audio}
Chenshuang Zhang, Chaoning Zhang, Sheng Zheng, Mengchun Zhang, Maryam Qamar,
  Sung-Ho Bae, and In~So Kweon.
\newblock A survey on audio diffusion models: Text to speech synthesis and
  enhancement in generative ai.
\newblock \emph{arXiv preprint arXiv:2303.13336}, 2023{\natexlab{d}}.

\bibitem[Li et~al.(2023)Li, Zhang, Waghwase, Lee, Rameau, Yang, Bae, and
  Hong]{li2023generative}
Chenghao Li, Chaoning Zhang, Atish Waghwase, Lik-Hang Lee, Francois Rameau,
  Yang Yang, Sung-Ho Bae, and Choong~Seon Hong.
\newblock Generative ai meets 3d: A survey on text-to-3d in aigc era.
\newblock \emph{arXiv preprint arXiv:2305.06131}, 2023.

\bibitem[Zhang et~al.(2022{\natexlab{a}})Zhang, Zhang, Song, Yi, Zhang, and
  Kweon]{zhang2022survey}
Chaoning Zhang, Chenshuang Zhang, Junha Song, John Seon~Keun Yi, Kang Zhang,
  and In~So Kweon.
\newblock A survey on masked autoencoder for self-supervised learning in vision
  and beyond.
\newblock \emph{arXiv preprint arXiv:2208.00173}, 2022{\natexlab{a}}.

\bibitem[Kirillov et~al.(2023)Kirillov, Mintun, Ravi, Mao, Rolland, Gustafson,
  Xiao, Whitehead, Berg, Lo, et~al.]{kirillov2023segment}
Alexander Kirillov, Eric Mintun, Nikhila Ravi, Hanzi Mao, Chloe Rolland, Laura
  Gustafson, Tete Xiao, Spencer Whitehead, Alexander~C Berg, Wan-Yen Lo, et~al.
\newblock Segment anything.
\newblock \emph{arXiv preprint arXiv:2304.02643}, 2023.

\bibitem[Zhang et~al.(2023{\natexlab{e}})Zhang, Zheng, Li, Qiao, Kang, Shan,
  Zhang, Qin, Rameau, Bae, et~al.]{zhang2023survey}
Chaoning Zhang, Sheng Zheng, Chenghao Li, Yu~Qiao, Taegoo Kang, Xinru Shan,
  Chenshuang Zhang, Caiyan Qin, Francois Rameau, Sung-Ho Bae, et~al.
\newblock A survey on segment anything model (sam): Vision foundation model
  meets prompt engineering.
\newblock \emph{arXiv preprint arXiv:2306.06211}, 2023{\natexlab{e}}.

\bibitem[Kevmo(2023)]{kevmo2023magiccopy}
Kevmo.
\newblock magic-copy, 2023.
\newblock URL \url{https://github.com/kevmo314/magic-copy}.
\newblock GitHub repository.

\bibitem[Adamdad(2023)]{adamdad2023anthing}
Adamdad.
\newblock Anything 3d, 2023.
\newblock URL \url{https://github.com/Anything-of-anything/Anything-3D}.
\newblock GitHub repository.

\bibitem[Chen(2023)]{chen2023box}
Yukang Chen.
\newblock 3d box segment anything, 2023.
\newblock URL \url{https://github.com/dvlab-research/3D-Box-Segment-Anything}.
\newblock GitHub repository.

\bibitem[Qiao et~al.(2023)Qiao, Zhang, Kang, Kim, Tariq, Zhang, and
  Hong]{qiao2023robustness}
Yu~Qiao, Chaoning Zhang, Taegoo Kang, Donghun Kim, Shehbaz Tariq, Chenshuang
  Zhang, and Choong~Seon Hong.
\newblock Robustness of sam: Segment anything under corruptions and beyond.
\newblock \emph{arXiv preprint arXiv:2306.07713}, 2023.

\bibitem[Zhang et~al.(2023{\natexlab{f}})Zhang, Zhou, Liang, and
  Chen]{zhang2023input}
Yizhe Zhang, Tao Zhou, Peixian Liang, and Danny~Z Chen.
\newblock Input augmentation with sam: Boosting medical image segmentation with
  segmentation foundation model.
\newblock \emph{arXiv preprint arXiv:2304.11332}, 2023{\natexlab{f}}.

\bibitem[Tang et~al.(2023)Tang, Xiao, and Li]{tang2023can}
Lv~Tang, Haoke Xiao, and Bo~Li.
\newblock Can sam segment anything? when sam meets camouflaged object
  detection.
\newblock \emph{arXiv preprint arXiv:2304.04709}, 2023.

\bibitem[Zhang et~al.(2023{\natexlab{g}})Zhang, Zhang, Kang, Kim, Bae, and
  Kweon]{zhang2023attack}
Chenshuang Zhang, Chaoning Zhang, Taegoo Kang, Donghun Kim, Sung-Ho Bae, and
  In~So Kweon.
\newblock Attack-sam: Towards evaluating adversarial robustness of segment
  anything model.
\newblock \emph{arXiv preprint arXiv:2305.00866}, 2023{\natexlab{g}}.

\bibitem[Moosavi-Dezfooli et~al.(2017{\natexlab{a}})Moosavi-Dezfooli, Fawzi,
  Fawzi, and Frossard]{moosavi2017universal}
Seyed-Mohsen Moosavi-Dezfooli, Alhussein Fawzi, Omar Fawzi, and Pascal
  Frossard.
\newblock Universal adversarial perturbations.
\newblock In \emph{CVPR}, 2017{\natexlab{a}}.

\bibitem[Moosavi-Dezfooli et~al.(2016)Moosavi-Dezfooli, Fawzi, and
  Frossard]{moosavi2016deepfool}
Seyed-Mohsen Moosavi-Dezfooli, Alhussein Fawzi, and Pascal Frossard.
\newblock Deepfool: a simple and accurate method to fool deep neural networks.
\newblock In \emph{CVPR}, 2016.

\bibitem[Madry et~al.(2018)Madry, Makelov, Schmidt, Tsipras, and
  Vladu]{madry2017towards}
Aleksander Madry, Aleksandar Makelov, Ludwig Schmidt, Dimitris Tsipras, and
  Adrian Vladu.
\newblock Towards deep learning models resistant to adversarial attacks.
\newblock In \emph{ICLR}, 2018.

\bibitem[Mopuri et~al.(2017{\natexlab{a}})Mopuri, Garg, and
  Babu]{Mopuri2017datafree}
Konda~Reddy Mopuri, Utsav Garg, and R.~Venkatesh Babu.
\newblock Fast feature fool: A data independent approach to universal
  adversarial perturbations.
\newblock In \emph{BMVC}, 2017{\natexlab{a}}.

\bibitem[Ma and Wang(2023)]{ma2023segment}
Jun Ma and Bo~Wang.
\newblock Segment anything in medical images.
\newblock \emph{arXiv preprint arXiv:2304.12306}, 2023.

\bibitem[Han et~al.(2023)Han, Zhang, Qiao, Qamar, Jung, Lee, Bae, and
  Hong]{han2023segment}
Dongsheng Han, Chaoning Zhang, Yu~Qiao, Maryam Qamar, Yuna Jung, SeungKyu Lee,
  Sung-Ho Bae, and Choong~Seon Hong.
\newblock Segment anything model (sam) meets glass: Mirror and transparent
  objects cannot be easily detected.
\newblock \emph{arXiv preprint}, 2023.

\bibitem[Shen et~al.(2023)Shen, Yang, and Wang]{shen2023anything}
Qiuhong Shen, Xingyi Yang, and Xinchao Wang.
\newblock Anything-3d: Towards single-view anything reconstruction in the wild.
\newblock \emph{arXiv preprint arXiv:2304.10261}, 2023.

\bibitem[Kang et~al.(2022)Kang, Min, and Hwang]{kang2022any}
Minki Kang, Dongchan Min, and Sung~Ju Hwang.
\newblock Any-speaker adaptive text-to-speech synthesis with diffusion models.
\newblock \emph{arXiv preprint arXiv:2211.09383}, 2022.

\bibitem[IDEA-Research(2023)]{GroundedSegmentAnything2023}
IDEA-Research.
\newblock Grounded segment anything, 2023.
\newblock URL \url{https://github.com/IDEA-Research/Grounded-Segment-Anything}.
\newblock GitHub repository.

\bibitem[Chen et~al.(2023)Chen, Yang, and Zhang]{chen2023semantic}
Jiaqi Chen, Zeyu Yang, and Li~Zhang.
\newblock Semantic-segment-anything, 2023.
\newblock URL \url{https://github.com/fudan-zvg/Semantic-Segment-Anything}.
\newblock GitHub repository.

\bibitem[Park(2023)]{park2023segment}
Curt Park.
\newblock segment anything with clip, 2023.
\newblock URL \url{https://github.com/Curt-Park/segment-anything-with-clip}.
\newblock GitHub repository.

\bibitem[Li et~al.(2022)Li, Li, Xiong, and Hoi]{li2022blip}
Junnan Li, Dongxu Li, Caiming Xiong, and Steven Hoi.
\newblock Blip: Bootstrapping language-image pre-training for unified
  vision-language understanding and generation.
\newblock In \emph{ICML}, pages 12888--12900. PMLR, 2022.

\bibitem[Radford et~al.(2021)Radford, Kim, Hallacy, Ramesh, Goh, Agarwal,
  Sastry, Askell, Mishkin, Clark, et~al.]{radford2021learning}
Alec Radford, Jong~Wook Kim, Chris Hallacy, Aditya Ramesh, Gabriel Goh,
  Sandhini Agarwal, Girish Sastry, Amanda Askell, Pamela Mishkin, Jack Clark,
  et~al.
\newblock Learning transferable visual models from natural language
  supervision.
\newblock In \emph{ICML}, 2021.

\bibitem[Rombach et~al.(2022)Rombach, Blattmann, Lorenz, Esser, and
  Ommer]{rombach2022high}
Robin Rombach, Andreas Blattmann, Dominik Lorenz, Patrick Esser, and Bj{\"o}rn
  Ommer.
\newblock High-resolution image synthesis with latent diffusion models.
\newblock In \emph{Proceedings of the IEEE/CVF Conference on Computer Vision
  and Pattern Recognition}, pages 10684--10695, 2022.

\bibitem[Yu et~al.(2023)Yu, Feng, Feng, Liu, Jin, Zeng, and
  Chen]{yu2023inpaint}
Tao Yu, Runseng Feng, Ruoyu Feng, Jinming Liu, Xin Jin, Wenjun Zeng, and Zhibo
  Chen.
\newblock Inpaint anything: Segment anything meets image inpainting.
\newblock \emph{arXiv preprint arXiv:2304.06790}, 2023.

\bibitem[Yang et~al.(2023)Yang, Gao, Li, Gao, Wang, and Zheng]{yang2023track}
Jinyu Yang, Mingqi Gao, Zhe Li, Shang Gao, Fangjing Wang, and Feng Zheng.
\newblock Track anything: Segment anything meets videos.
\newblock \emph{arXiv preprint arXiv:2304.11968}, 2023.

\bibitem[Zxyang(2023)]{z-x-yang_2023}
Zxyang.
\newblock Segment and track anything, 2023.
\newblock URL \url{https://github.com/z-x-yang/Segment-and-Track-Anything}.
\newblock GitHub repository.

\bibitem[Zhang et~al.(2023{\natexlab{h}})Zhang, Han, Qiao, Kim, Bae, Lee, and
  Hong]{zhang2023faster}
Chaoning Zhang, Dongshen Han, Yu~Qiao, Jung~Uk Kim, Sung~Ho Bae, Seungkyu Lee,
  and Choong~Seon Hong.
\newblock Faster segment anything: Towards lightweight sam for mobile
  applications.
\newblock \emph{arXiv preprint arXiv:2306.14289}, 2023{\natexlab{h}}.

\bibitem[Zhang et~al.(2023{\natexlab{i}})Zhang, Qiao, Tariq, Zheng, Zhang, Li,
  Shin, and Hong]{zhang2023understanding}
Chaoning Zhang, Yu~Qiao, Shehbaz Tariq, Sheng Zheng, Chenshuang Zhang, Chenghao
  Li, Hyundong Shin, and Choong~Seon Hong.
\newblock Understanding segment anything model: Sam is biased towards texture
  rather than shape.
\newblock 2023{\natexlab{i}}.

\bibitem[Zhang et~al.(2023{\natexlab{j}})Zhang, Zhang, Kang, Kim, Bae, and
  Kweon]{zhang2023attacksam}
Chenshuang Zhang, Chaoning Zhang, Taegoo Kang, Donghun Kim, Sung-Ho Bae, and
  In~So Kweon.
\newblock Attack-sam: Towards evaluating adversarial robustness of segment
  anything model.
\newblock \emph{arXiv preprint}, 2023{\natexlab{j}}.

\bibitem[Khrulkov and Oseledets(2018)]{khrulkov2018art}
Valentin Khrulkov and Ivan Oseledets.
\newblock Art of singular vectors and universal adversarial perturbations.
\newblock In \emph{CVPR}, 2018.

\bibitem[Mopuri et~al.(2018{\natexlab{a}})Mopuri, Ojha, Garg, and
  Babu]{mopuri2018nag}
Konda~Reddy Mopuri, Utkarsh Ojha, Utsav Garg, and R.~Venkatesh Babu.
\newblock Nag: Network for adversary generation.
\newblock In \emph{CVPR}, 2018{\natexlab{a}}.

\bibitem[Perolat et~al.(2018)Perolat, Malinowski, Piot, and
  Pietquin]{perolat2018playing}
Julien Perolat, Mateusz Malinowski, Bilal Piot, and Olivier Pietquin.
\newblock Playing the game of universal adversarial perturbations.
\newblock \emph{arXiv preprint arXiv:1809.07802}, 2018.

\bibitem[Mopuri et~al.(2017{\natexlab{b}})Mopuri, Garg, and
  Babu]{mopuri2017fast}
Konda~Reddy Mopuri, Utsav Garg, and R~Venkatesh Babu.
\newblock Fast feature fool: A data independent approach to universal
  adversarial perturbations.
\newblock \emph{arXiv preprint arXiv:1707.05572}, 2017{\natexlab{b}}.

\bibitem[Mopuri et~al.(2018{\natexlab{b}})Mopuri, Ganeshan, and
  Radhakrishnan]{mopuri2018generalizable}
Konda~Reddy Mopuri, Aditya Ganeshan, and Venkatesh~Babu Radhakrishnan.
\newblock Generalizable data-free objective for crafting universal adversarial
  perturbations.
\newblock \emph{TPAMI}, 2018{\natexlab{b}}.

\bibitem[Zhang et~al.(2020{\natexlab{a}})Zhang, Benz, Imtiaz, and
  Kweon]{zhang2019cd-uap}
Chaoning Zhang, Philipp Benz, Tooba Imtiaz, and In-So Kweon.
\newblock Cd-uap: Class discriminative universal adversarial perturbation.
\newblock In \emph{AAAI}, 2020{\natexlab{a}}.

\bibitem[Benz et~al.(2020)Benz, Zhang, Imtiaz, and Kweon]{benz2020double}
Philipp Benz, Chaoning Zhang, Tooba Imtiaz, and In~So Kweon.
\newblock Double targeted universal adversarial perturbations.
\newblock In \emph{ACCV}, 2020.

\bibitem[Kingma and Ba(2015)]{kingma2014adam}
Diederik~P Kingma and Jimmy Ba.
\newblock Adam: A method for stochastic optimization.
\newblock In \emph{ICLR}, 2015.

\bibitem[Goodfellow et~al.(2015)Goodfellow, Shlens, and
  Szegedy]{goodfellow2014explaining}
Ian~J Goodfellow, Jonathon Shlens, and Christian Szegedy.
\newblock Explaining and harnessing adversarial examples.
\newblock In \emph{ICLR}, 2015.

\bibitem[Zhang et~al.(2020{\natexlab{b}})Zhang, Benz, Imtiaz, and
  Kweon]{zhang2020understanding}
Chaoning Zhang, Philipp Benz, Tooba Imtiaz, and In-So Kweon.
\newblock Understanding adversarial examples from the mutual influence of
  images and perturbations.
\newblock In \emph{CVPR}, 2020{\natexlab{b}}.

\bibitem[Zhang et~al.(2021)Zhang, Benz, Karjauv, and Kweon]{zhang2021data}
Chaoning Zhang, Philipp Benz, Adil Karjauv, and In~So Kweon.
\newblock Data-free universal adversarial perturbation and black-box attack.
\newblock In \emph{ICCV}, 2021.

\bibitem[Schroff et~al.(2015)Schroff, Kalenichenko, and
  Philbin]{Schroff2015FaceNetAU}
Florian Schroff, Dmitry Kalenichenko, and James Philbin.
\newblock Facenet: A unified embedding for face recognition and clustering.
\newblock \emph{2015 IEEE Conference on Computer Vision and Pattern Recognition
  (CVPR)}, 2015.

\bibitem[Wang and Gupta(2015)]{wang2015unsupervised}
Xiaolong Wang and Abhinav Gupta.
\newblock Unsupervised learning of visual representations using videos.
\newblock In \emph{ICCV}, 2015.

\bibitem[Sohn(2016)]{sohn2016improved}
Kihyuk Sohn.
\newblock Improved deep metric learning with multi-class n-pair loss objective.
\newblock In \emph{NeurIPS}, 2016.

\bibitem[Misra et~al.(2016)Misra, Zitnick, and Hebert]{misra2016shuffle}
Ishan Misra, C~Lawrence Zitnick, and Martial Hebert.
\newblock Shuffle and learn: unsupervised learning using temporal order
  verification.
\newblock In \emph{ECCV}, 2016.

\bibitem[Federici et~al.(2020)Federici, Dutta, Forré, Kushman, and
  Akata]{Federici2020Learning}
Marco Federici, Anjan Dutta, Patrick Forré, Nate Kushman, and Zeynep Akata.
\newblock Learning robust representations via multi-view information
  bottleneck.
\newblock In \emph{ICLR}, 2020.
\newblock URL \url{https://openreview.net/forum?id=B1xwcyHFDr}.

\bibitem[Hadsell et~al.(2006)Hadsell, Chopra, and
  LeCun]{hadsell2006dimensionality}
Raia Hadsell, Sumit Chopra, and Yann LeCun.
\newblock Dimensionality reduction by learning an invariant mapping.
\newblock In \emph{CVPR}, 2006.

\bibitem[Hermans et~al.(2017)Hermans, Beyer, and Leibe]{hermans2017defense}
Alexander Hermans, Lucas Beyer, and Bastian Leibe.
\newblock In defense of the triplet loss for person re-identification.
\newblock \emph{arXiv preprint arXiv:1703.07737}, 2017.

\bibitem[Wu et~al.(2018)Wu, Xiong, Yu, and Lin]{wu2018unsupervised}
Zhirong Wu, Yuanjun Xiong, Stella~X Yu, and Dahua Lin.
\newblock Unsupervised feature learning via non-parametric instance
  discrimination.
\newblock In \emph{CVPR}, 2018.

\bibitem[Oord et~al.(2018)Oord, Li, and Vinyals]{oord2018representation}
Aaron van~den Oord, Yazhe Li, and Oriol Vinyals.
\newblock Representation learning with contrastive predictive coding.
\newblock \emph{arXiv preprint arXiv:1807.03748}, 2018.

\bibitem[Chen et~al.(2020{\natexlab{a}})Chen, Kornblith, Norouzi, and
  Hinton]{chen2020simple}
Ting Chen, Simon Kornblith, Mohammad Norouzi, and Geoffrey Hinton.
\newblock A simple framework for contrastive learning of visual
  representations.
\newblock In \emph{ICML}, 2020{\natexlab{a}}.

\bibitem[He et~al.(2020)He, Fan, Wu, Xie, and Girshick]{he2020momentum}
Kaiming He, Haoqi Fan, Yuxin Wu, Saining Xie, and Ross Girshick.
\newblock Momentum contrast for unsupervised visual representation learning.
\newblock In \emph{CVPR}, 2020.

\bibitem[Chen et~al.(2020{\natexlab{b}})Chen, Fan, Girshick, and
  He]{chen2020mocov2}
Xinlei Chen, Haoqi Fan, Ross Girshick, and Kaiming He.
\newblock Improved baselines with momentum contrastive learning.
\newblock \emph{arXiv preprint arXiv:2003.04297}, 2020{\natexlab{b}}.

\bibitem[He et~al.(2022)He, Chen, Xie, Li, Doll{\'a}r, and
  Girshick]{he2022masked}
Kaiming He, Xinlei Chen, Saining Xie, Yanghao Li, Piotr Doll{\'a}r, and Ross
  Girshick.
\newblock Masked autoencoders are scalable vision learners.
\newblock In \emph{CVPR}, 2022.

\bibitem[Moosavi-Dezfooli et~al.(2017{\natexlab{b}})Moosavi-Dezfooli, Fawzi,
  Fawzi, Frossard, and Soatto]{moosavi2017analysis}
Seyed-Mohsen Moosavi-Dezfooli, Alhussein Fawzi, Omar Fawzi, Pascal Frossard,
  and Stefano Soatto.
\newblock Analysis of universal adversarial perturbations.
\newblock \emph{arXiv preprint arXiv:1705.09554}, 2017{\natexlab{b}}.

\bibitem[Wang and Liu(2021)]{Wang_2021_CVPR}
Feng Wang and Huaping Liu.
\newblock Understanding the behaviour of contrastive loss.
\newblock In \emph{CVPR}, 2021.

\bibitem[Zhang et~al.(2022{\natexlab{b}})Zhang, Zhang, Pham, Yoo, and
  Kweon]{zhang2022dual}
Chaoning Zhang, Kang Zhang, Trung~X. Pham, Changdong Yoo, and In-So Kweon.
\newblock Dual temperature helps contrastive learning without many negative
  samples: Towards understanding and simplifying moco.
\newblock In \emph{CVPR}, 2022{\natexlab{b}}.

\end{thebibliography}

\end{document}